\documentclass[sigconf]{acmart}
\acmSubmissionID{1570}

\usepackage[american]{babel} 
\usepackage{microtype} 

\usepackage{booktabs} 
\usepackage{color}
\usepackage{balance}
\usepackage{bm}

\usepackage[ruled]{algorithm2e} 

\SetAlFnt{\small}
\SetAlCapFnt{\small}
\SetAlCapNameFnt{\small}
\SetAlCapHSkip{0pt}

\copyrightyear{2020} 
\acmYear{2020} 
\setcopyright{acmcopyright}\acmConference[MM '20]{Proceedings of the 28th ACM International Conference on Multimedia}{October 12--16, 2020}{Seattle, WA, USA}
\acmBooktitle{Proceedings of the 28th ACM International Conference on Multimedia (MM '20), October 12--16, 2020, Seattle, WA, USA}
\acmPrice{15.00}
\acmDOI{10.1145/3394171.3413684}
\acmISBN{978-1-4503-7988-5/20/10}

\newcommand{\real}{{\mathbb{R}}}
\newcommand{\synS}{\mathcal{S}_{syn}}
\newcommand{\dfmS}{\mathcal{S}_{dfm}}
\newcommand{\hdS}{\mathcal{S}}

\begin{document}
\fancyhead{}	

\title{DeepFacePencil: Creating Face Images from Freehand Sketches}

\author{Yuhang Li}
\email{lyh9001@mail.ustc.edu.cn}
\affiliation{%
\institution{NEL-BITA, University of Science and Technology of China}
}

\author{Xuejin Chen}
\authornote{Xuejin Chen is the corresponding author.}
\email{xjchen99@ustc.edu.cn}
\affiliation{%
\institution{NEL-BITA, University of Science and Technology of China}
}

\author{Binxin Yang} 
\email{tennyson@mail.ustc.edu.cn}
\affiliation{%
\institution{NEL-BITA, University of Science and Technology of China}
}

\author{Zihan Chen} 
\email{zhchen25@mail.ustc.edu.cn}
\affiliation{%
	\institution{NEL-BITA, University of Science and Technology of China}
}

\author{Zhihua Cheng} 
\email{czh666@mail.ustc.edu.cn}
\affiliation{%
	\institution{NEL-BITA, University of Science and Technology of China}
}

\author{Zheng-Jun Zha} 
\email{zhazj@ustc.edu.cn}
\affiliation{%
\institution{NEL-BITA, University of Science and Technology of China}
}

\begin{abstract}
In this paper, we explore the task of generating photo-realistic face images from hand-drawn sketches. Existing image-to-image translation methods require a large-scale dataset of paired sketches and images for supervision. They typically utilize synthesized edge maps of face images as training data. However, these synthesized edge maps strictly align with the edges of the corresponding face images, which limit their generalization ability to real hand-drawn sketches with vast stroke diversity. To address this problem, we propose DeepFacePencil, an effective tool that is able to generate photo-realistic face images from hand-drawn sketches, based on a novel dual generator image translation network during training. A novel spatial attention pooling (SAP) is designed to adaptively handle stroke distortions which are spatially varying to support various stroke styles and different level of details. We conduct extensive experiments and the results demonstrate the superiority of our model over existing methods on both image quality and model generalization to hand-drawn sketches.
\end{abstract}

%



\maketitle

\section{Introduction}

Flexibly creating new content is one of the most important goals in both computer graphics and computer-human interaction. While sketching is an efficient and natural way for common users to express their ideas for designing and editing new content, sketch-based interaction techniques have been extensively studied~\cite{SutherlandSketchPad64,Zeleznik-Sketch96,Igarashi-teddy99,Chen_sketchingreality08,Sketch2Photo}. 
Imagery content is the most ubiquitous media with a large variety of display devices everywhere in our daily life. 
Creating new imagery content is one way to show people's creativity and communicate smart ideas. In this paper, we target portrait imagery, which is inextricably bound to our life, and present a sketch-based system, \emph{DeepFacePencil}, which allows common users to create new face imagery by specifying the desired facial shapes via freehand sketching. 

Deep learning techniques have brought significant improvements on the realism of virtual images. 
Recently, a large amount of studies have been conducted on general image-to-image translation which aims to translate an image from one domain to a corresponding image in another domain, while preserving the same content, such as structure, scenes or objects~\cite{pix2pix,pix2pixHD,CycleGANs,DiscoGANs, DualGANs,BicycleGANs}. Treating sketches as the source domain and realistic face images as the target domain, this task is a typical image-to-image translation problem.
However, exiting image-to-image translation techniques are not off-the-shelf for this task due to the underlying challenges: \emph{data scarcity} in the sketch domain and \emph{ambiguity} in freehand sketches.

\begin{figure}
	\centering
	\includegraphics[width=\columnwidth]{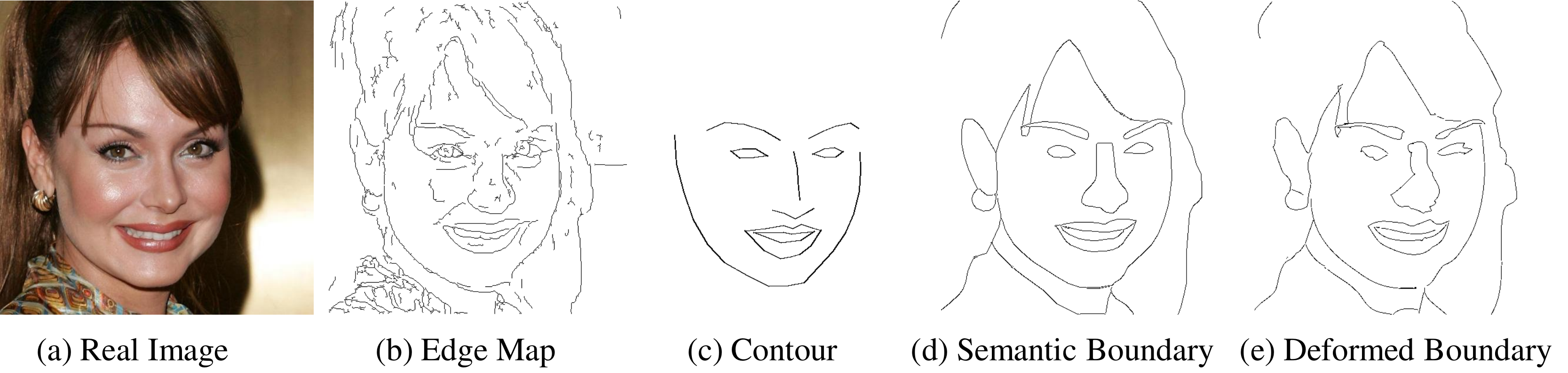}
	\caption{From a real face image, different styles of synthesized sketches can be generated by edge detection, contour alignment, or semantic segmentation.}
	\label{fig:sketch_data}
\end{figure}

Since there exists no large-scale dataset of paired sketch and face images and collecting hand-drawn sketches is time-consuming, existing methods~\cite{pix2pix, pix2pixHD, Lines2Face} utilize edge maps or contours of real face images as training data when applied on the sketch-to-face task. Figure~\ref{fig:sketch_data} shows multiple styles of synthesized edge maps, face contours or semantic boundaries. 
These synthesized edge maps or contours enable existing models to be trained in a supervised manner and obtain plausible results on synthesized edge maps or contours. 
However, models trained on synthesized data are not able to achieve satisfactory results on hand-drawn sketches, specially on those drawn by users without considerable drawing skills. 

Since strokes in edge maps and contours align perfectly with edges of the corresponding real images, models trained on edge-aligned data tend to generate unreal shapes of facial parts following the inaccurate strokes when the input sketch is poorly drawn. Hence, for an imperfect hand-drawn sketch, there is a trade-off between \textit{the realism} of the synthesized image and \textit{the conformance} between the input sketch and the edges of the synthesized image.
Models with strong edge alignment fail to be generalized to hand-drawn sketches that contains many imperfect strokes.

Moreover, we observe that the balance between image realism and shape conformance mentioned above varies from one position to another across a synthesized image. In a portrait sketch, some facial parts might be well drawn while the others are not. For the well-drawn facial parts, the balance is supposed to move towards the conformance in order to ensure those parts in the synthesized image depicting the user's imagination. On the other hand, for the poorly-drawn parts, the image realism should be emphasized by not strictly following the irregular shapes and strokes.

Based on the discussion above, we propose a novel sketch-based face image synthesis framework that is robust to hand-drawn sketches. A new module, named spatial attention pooling (SAP), is designed to adjust the spatially varying balance between \textit{realism} and \textit{conformance} adaptively across an image. In order to break the strict alignment between sketches and real images, our SAP relaxes strokes with one-pixel width to multiple-pixel widths using pooling operators. A larger width of a stroke, which is controlled by the kernel size of a pooling operator, indicates less restriction between this stroke and the corresponding edge in the synthesized image. However, the kernel size is not trainable via back propagation. Hence, for an input sketch, multiple branches of pooling operators with different kernel sizes are added in SAP to get multiple relaxed sketches with different widths. The relaxed sketches are then fused by a spatial attention layer which adjusts the balance of \textit{realism} and \textit{conformance}. For different location in a portrait sketch, the spatial attention layer assigns high attention to the relaxed sketch with large width if this position requires more \textit{realism} than \textit{conformance}.
In order to more effectively train the SAP module, we propose a dual-generator framework to enforce the embedded features after SAP from imperfect sketches to be consistent with the features of well-synthesized boundaries from real images. This dual-generator training framework promotes the SAP to perceive the imperfectness of local strokes and rectify the synthesized face regions from the imperfect stroke to realistic image domain.

In summary, our contribution in this paper is three-fold.
\begin{itemize}
\item Based on comprehensive analysis on the edge alignment issue in image translation frameworks, we propose a sketch-to-face translation system that is robust to hand-drawn sketches with various drawing skills. 
\item A novel deep neural network module for sketch, named \emph{spatial attention pooling}, is designed to adaptively adjust the spatially varying balance between the realism of the synthesized image and the conformance between the input sketch and the synthesized image.
\item We propose a dual-generator framework which effectively trains the SAP to sense the distortion of imperfect strokes and rectify the embedding features to realistic face domain. 
\end{itemize}

\begin{figure*}[ht]
	\includegraphics[width=0.9\textwidth]{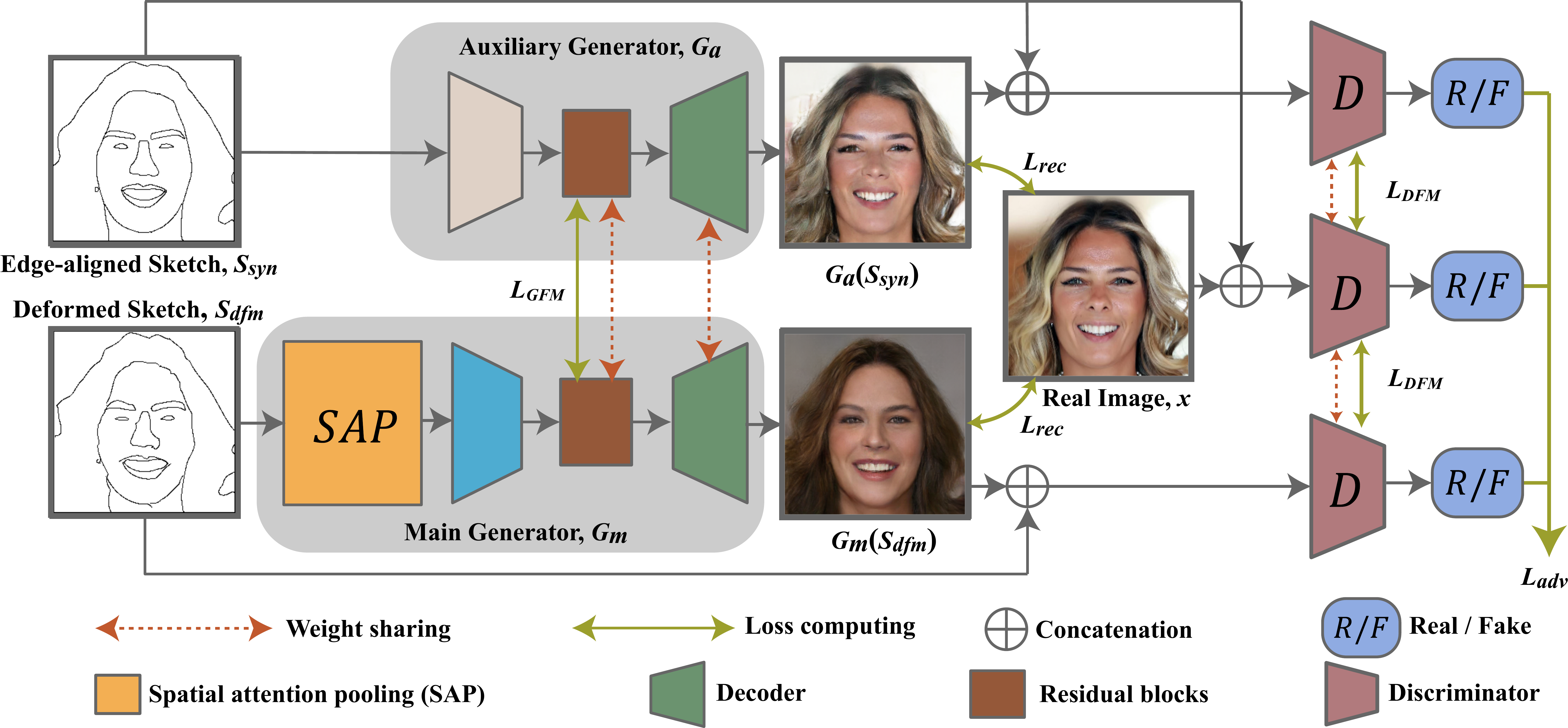}
	\caption{The architecture of our dual-generator model with spatial attention pooling (SAP) for deformed sketches. In order to train a face image generator $G_m$ for hand-drawn sketches, we synthesize deformed sketches $\dfmS$ from an edge-aligned sketch $\synS$ and design a spatially attention pooling module to extract shape and structure features from distorted sketches. The dual generators $G_m$ and $G_a$ are trained simultaneously with three discriminators in an adversarial manner. }
	\label{fig:architecture}
\end{figure*}

\section{Related Work}
Our method is related to studies on image-to-image translation, sketch-based image generation, as well as face image generation and editing.
In this section, we discuss the approaches that are most related to our work.

\subsection{Image-to-Image Translation}
Given an input image from one domain, an image-to-image translation model outputs a corresponding image from another domain and preserves the content in the input image. Existing image-to-image translation models are based on generative adversarial networks conditioned on images. Pix2pix~\cite{pix2pix} is the first general image-to-image translation model which can be applied to different scenarios according to paired training images, such as, semantic maps to natural images, day images to night images, image colorization, and edge maps to images. \cite{outdoor_scene} utilizes semantic label maps and attributes of outdoor scenes as input and generates the corresponding photo-realistic images. In order to model multi-modal distribution of output images, BicycleGAN~\cite{BicycleGANs} encourages the connection between the output and the latent code to be invertible. CycleGAN~\cite{CycleGANs}, DualGAN~\cite{DualGANs}, and DiscoGAN~\cite{DiscoGANs} propose unsupervised image translation models with a common idea named cycle consistency, which is borrowed from language translation literature. Pix2pixHD~\cite{pix2pixHD} is proposed as a high-resolution image-to-image translation model for generating photo-realistic image from semantic label maps using a coarse-to-fine generator and a multi-scale discriminator. It can also be applied to edge-to-photo generation when trained on paired edge maps and photos. However, the large gap between synthesized edge maps and hand-drawn sketches challenges the generalization of these models.

\subsection{Sketch-based Image Generation}
Sketch-based image generation is a hot topic in multimedia and computer graphics. Given a sketch which describes the desired scene layout with text labels for objects, traditional methods, such as Sketch2\-Photo \cite{Sketch2Photo} and PhotoSketcher~\cite{PhotoSketcher}, search image patches from a large-scale image dataset and fuse the retrieved image patches together according to the sketch. These methods are not able to ensure the global consistency of the resultant image and fail to generate totally new images. Nevertheless, it is challenging for these methods to ensure global consistency of the resultant images. Thus they frequently fail to generate totally new images.
After the breakthroughs made by deep neural networks (DNNs) in many image understanding tasks, a variety of DNN-based models have been proposed for sketch-based image generation. The general image-to-image translation models mentioned above can be easily extended to sketch-based image generation once sketches and their corresponding images are available as training data. Besides, a few other models are designed specially for sketch inputs. SketchyGAN~\cite{SketchyGAN} aims to generate real images from multi-class sketches. A novel neural network module, called mask residual unit (MRU), is proposed to improve the information flow by injecting the input image at multiple scales. Edge maps are extracted from real images and utilized as training sketches. However, the resultant images of SketchyGAN are still not satisfied. LinesToFacePhoto~\cite{Lines2Face} employs a conditional self-attention module to preserve the completeness of global facial structure in generated face images.
However, this model cannot be generalized to hand-drawn sketches directly due to distinct stroke characteristics.

\subsection{Face Image Generation and Editing}

Recently, studies on face image generation and editing have made tremendous progress. Generative adversarial network (GAN) \cite{GANs}, which generates images from noise, is widely used in a wide range of applications \cite{SRGANs, StackGANs, StackGANs++, 9156753, 8953294, 8985326}. Using GAN, realistic face images can be generated from noise vectors. DCGAN \cite{DCGANs} introduces a novel network to stabilize training of GAN.
PGGAN~\cite{PGGAN} utilizes a progressively growing architecture to generate high-resolution face images.
Inspired by style transfer literature, StyleGAN \cite{StyleGAN} introduces a novel generator which synthesizes plausible high-resolution face images and learns unsupervised separation of high-level attributes and stochastic variation in synthesized images. 
On the other side, a number of works focus on face image editing through different control information. StarGAN \cite{StarGAN-CVPR2018} designs a one-to-many translation framework which switches face attributes assigned by an attribute code. FaceShop \cite{Faceshop-Portenier-TOG18} and SC-FEGAN \cite{SC-FEGAN-Jo-ICCV2019} treat sketch-based face editing as a sketch-guided image inpainting problem where stroke colors are also applied as guidance information. 
In this work, we focus more on synthesizing realistic face images from imperfectly drawn sketches.


\section{Our Sketch-to-Photo Translation Network}
\label{sec:network}
 
%

The task of sketch-to-photo translation can be defined as looking for a generator $G$ so that the generated image $G(S)$ from a hand-drawn sketch $S$ looks realistic and keeps the shape characteristics of the input sketch. Existing image translation techniques train a neural network as the generator with paired of sketch and photo data $(\mathcal{S}, \mathcal{X})$. Due to the scarcity of real hand-drawn sketches, existing techniques synthesize sketches in a certain style to approximate the sketch set $\mathcal{S}$ from face image set $\mathcal{X}$ to train their generator in an adversarial manner.
The synthesized sketches $\mathcal{S}_{syn}$ are usually precisely aligned with the face images and present different distributions from hand-drawn sketches $\mathcal{S}$.
These models typically fail to generalize to hand-drawn sketches by common users. 

We propose a novel network architecture with a specially designed training strategy to improve the capability of the sketch-based image generator. Figure~\ref{fig:architecture} shows the overview of our method. In order to synthesize a set of sketches that has similar distribution with hand-drawn sketches $\hdS$, we deform the edge-aligned sketches $\synS$ to generate a set of deformed sketches $\dfmS$ to augment the training set.
We propose a novel framework using dual generators from the edge-aligned sketches $\synS$ and the deformed sketches $\dfmS$ respectively.
$G_m$ is the main generator trained with deformed sketches $\dfmS$, aiming to generate plausible photo-realistic face images from unseen hand-drawn sketches in the test stage. 
$G_a$ is an auxiliary generator trained with edge-aligned sketches whose goal is to guide $G_m$ to adaptively sense the line distortion in deformed sketches and rectify the distortion in feature space.
A spatial attention pooling module (SAP) is added before the encoder $E_m$ of $G_m$ to adjust the spatially varying balance between \textit{the realism} of generated images and \textit{the conformance} between the generated image and the input sketch.

\subsection{Synthesized Sketches and Deformation}
\label{subsec:algorithm_data}

Paired face sketch-photo dataset is required for supervised sketch-to-face translation methods. Since there exits no large-scale paired sketch-image dataset, the training sketches used by existing methods~\cite{pix2pix, Lines2Face} are generated from face image dataset, e.g. CelebA-HQ face dataset, using edge detection algorithm such as HED~\cite{HED}. However, the level of details in edge maps rely heavily on the value of a threshold in the edge detection algorithm. An edge map with a large threshold contains too many redundant edges while an edge map with a small threshold fails to preserve the entire global facial structure~\cite{Lines2Face}. Pix2pixHD~\cite{pix2pixHD} introduces another method to generate sketches from face images. Given a face image, the face landmarks are detected using an off-shelf landmark detection model. A new kind of sketch, denoted as \textit{face contour}, is obtained by connecting the landmarks in a fixed order. However, since the pre-defined face landmarks mainly depict the facial area while ignoring details, a sketch-to-face model trained by face contours fails to generalize to hand-drawn sketches with hair, beard, or ornaments . 

Based on the discussion above, we utilize a new kind of generated sketches with the assist of semantic maps.
The CelebAMask-HQ dataset~\cite{CelebAMask-HQ} provides a face semantic map for each face image in CelebA-HQ dataset. We use the boundaries of the semantic map as a synthesized sketch of the corresponding face image. Figure~\ref{fig:sketch_data} compares different styles of synthetic sketches, including an edge map (b), a face contour (c), region boundaries of semantic maps (d), and deformed boundary map (e) from the same real image (a).

\paragraph{Stroke Deformation}
A shortcoming of sketches that are generated from semantic boundary or edge detector is that the synthesized sketch lines are perfectly aligned to edges of the corresponding face images. In order to break the strict edge alignment between sketches and the corresponding real images and mimic stroke characteristics of hand-drawn sketches, we apply line deformation similar with FaceShop~\cite{Faceshop-Portenier-TOG18}. Specifically, we vectorize lines of each sketch using AutoTrace~\cite{AutoTrace}. Then offsets randomly selected from $[-d, d]^2$ are added to the control points and end points of the vectorized lines, where $d$ is the maximum offset to control the deformation degree. 
We set $d=11$ in our experiments unless specifically mentioned. We use the semantic boundary map as edge-aligned sketch $\synS$, and semantic boundary map with random deformation as deformed sketch $\dfmS$.

\begin{figure}
	\includegraphics[width=\columnwidth]{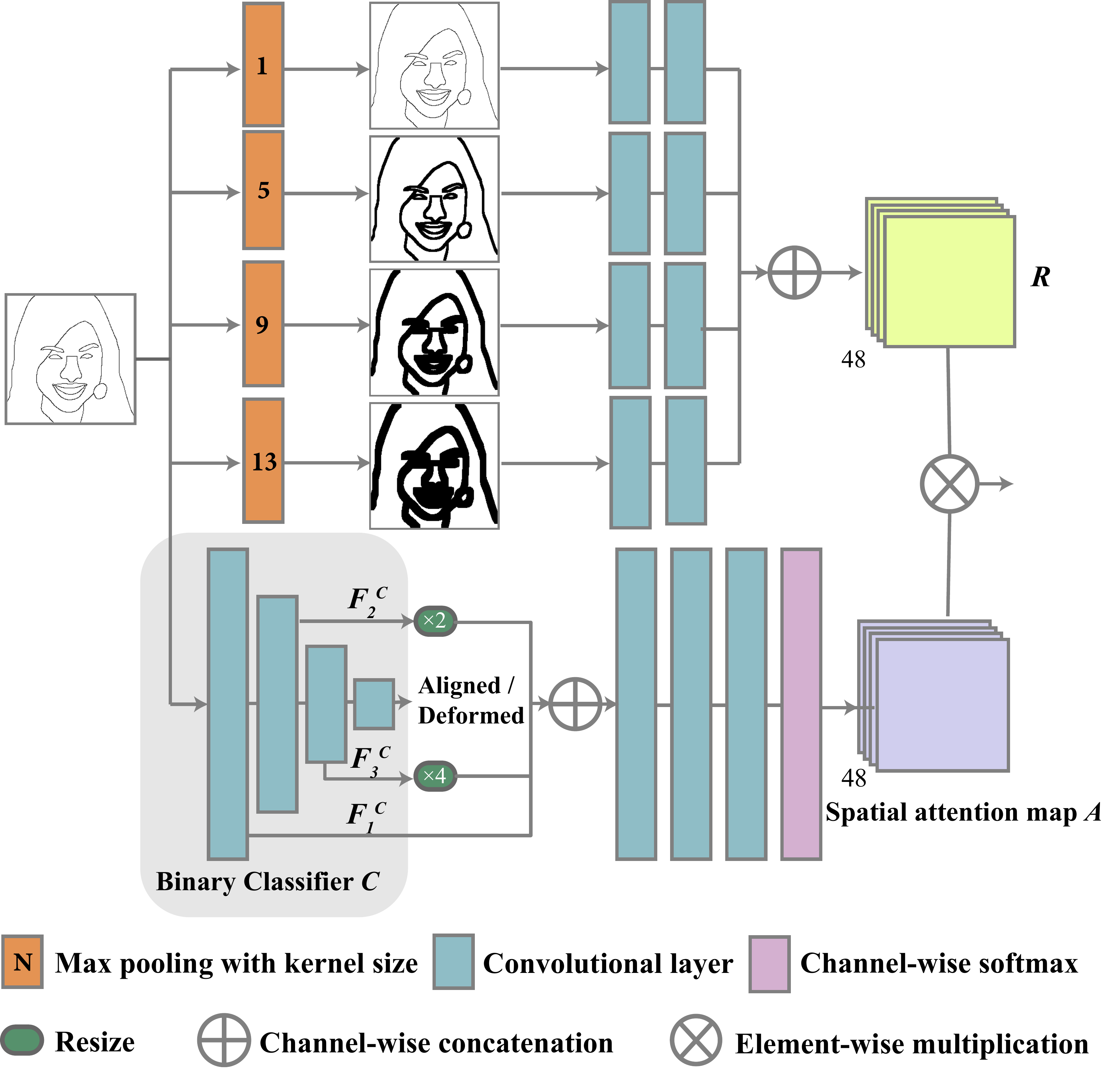}
	\caption{Network architecture of our spatial attention pooling module.}
	\label{fig:sap}
\end{figure}

\subsection{Spatial Attention Pooling}
\label{subsec:algorithm_sap}

A sketch-to-image model trained with edge-aligned sketch-image pairs tends to generate images whose edges strictly align with the strokes of the input sketch.
When an input hand-drawn sketch is not well-drawn, line distortions in the input sketch damage the quality of the generated face image. 
There is a trade-off between the realism of the generated face image and the conformance between the input sketch and the output face image. In order to alleviate the edge alignment between the input sketch and the output face image, we propose to relax thin strokes to a tolerance region with various width. A straightforward way is to smooth the strokes to multi-pixel width by image smoothing or dilation. 
However, the capacity of this hand-crafted way is limited, because the uniform smoothness for all positions of the whole sketch violates the unevenness of hand-drawn sketches on depicting different facial parts. We argue that the balance between the realism and the conformance differs from one position to another across the face image. 
Even in the same input sketch, the user might draw strokes in some parts (such as eyes and mouth) carefully with little distortions, while draw strokes in other parts (such as chin and hair) roughly. 
The balance moves towards conformance at a well-drawn part with realistic shape to meet the user's intention, while it moves towards realism at a poorly-drawn part to ensure the quality of the generated image.
Therefore, the relaxation degree should be spatially varying.

We propose a new module, called spatial attention pooling (SAP), to adaptively relax strokes in the input sketch to spatially varying tolerance regions. 
A stroke with a larger width indicates less restriction between this stroke and the corresponding edge in the synthesized image. The stroke widths are controlled by the kernel sizes of pooling operators. However, the kernel size of an pooling operator is not trainable using back propagation. Instead, we apply multiple pooling operators with different kernel sizes to get multiple relaxed sketches with different stroke widths. The relaxed sketches are then fused by a spatial attention layer which spatially adjusts the balance between \textit{realism} and \textit{conformance}. 
 
The architecture of SAP is shown in Figure~\ref{fig:sap}.
Given an input deformed sketch $\dfmS \in \real^{H\times W}$, we first pass it through $N_r$ pooling branches with different kernel sizes of $\{r_i\}_{i=1,\ldots, N_r}$ to get $P_{i}=Pooling(\dfmS;r_i)$. 
Then we utilize convolutional layers to extract feature maps of $P_i$ separately. These feature maps are concatenated to get a relaxed representation of $\dfmS$, denoted as $R$:
\begin{equation}
R=Concat\big(Conv_1(P_1), Conv_2(P_2),..., Conv_{N_r}(P_{N_r})\big),
\end{equation}
where $Conv_i()$ indicates convolutional layers for each pooling branch, and $Concat$ is a channel-wise concatenation operator.

In order to control the relax degrees of all positions, we compute a spatial attention map $A$ to assign different attention weights to $R$. 
A stroke with more distortion is supposed to be assigned with larger relax degree. 
Hence, $A$ is supposed to adaptively pay more attention (a larger weight) to a $Conv_i(P_i)$ that comes from a larger pooling kernel size in the areas with more distortions.
In order to perceive the distortion degree of different regions in a hand-drawn sketch, we employ a pre-trained binary classifier $C$ with three convolutional layers to distinguish edge-aligned $\synS$ sketches from deformed sketches $\dfmS$. We utilize this classifier to extract features of the input sketch $\dfmS$ to get ${F^{C}_i(\dfmS), i=1,2,3}$, where $F^{C}_i()$ denotes the feature maps extracted by the $i^{th}$ convolution layers of $C$. These feature maps emphasize the differences between edge-aligned sketches and deformed sketches. We resize and concatenate these feature maps, and pass them through three convolutional layers to get the spatial attention map:
\begin{equation}
A=softmax\Big(Conv\big([F^{C}_1, Up_2(F^{C}_2), Up_4(F^{C}_3)]\big)\Big), 
\end{equation}
where $Up_2$ and $Up_4$ indicates $2\times$ and $4\times$ upsampling, $Conv()$ indicates three cascaded convoutional layers, and $softmax()$ is a softmax layer computed over channels to ensuring that for each position of $A$, the sum of weights of all channels equals to $1$.

At last, the output SAP is computed as:
\begin{equation}
SAP(\dfmS)=A*R,
\end{equation}
where $*$ is element-wise multiplication.

%

\subsection{Training with Dual Generators}
\label{subsec:algorithm_loss}


The goal of SAP is to sense stroke distortions in different regions of a hand-drawn sketch and then relax the edge alignment accordingly in the synthesized face image.
However, directly adding the SAP module at the front end of a general image translation network does not guarantee that the SAP can be effectively trained under the loss with regard to the synthesized image only. 
Following a general image translation network architecture, the realism of synthesized images relies on the coincidence of the embedded structure from the input sketch and real face images. 
In our dual-generator framework shown in Figure \ref{fig:architecture}, we enforce the embedded features from the deformed sketch with SAP to be consistent with the synthesized edge-aligned sketches by introducing a generator feature matching loss $\mathcal{L}_{GFM}$ between the embedded features from the deformed sketch $\synS$ and the edge-aligned sketch $\dfmS$. 
The main generator $G_m$ for deformed sketches, the auxiliary generator $G_a$ for edge-aligned sketches, and the discriminator $D$ are trained with the following objectives.

\paragraph{Reconstruction Loss}
For either generator, a reconstruction loss is applied to guide the generated image close to its corresponding real image $x$.
\begin{equation}
\label{eqn:loss_rec}
\begin{aligned}
\mathcal{L}_{rec}(G_a) &= \mathbb{E}\|G_a(\synS) - x\|_1 \\
\mathcal{L}_{rec}(G_m) &= \mathbb{E}\|G_m(\dfmS) - x\|_1.
\end{aligned}
\end{equation}

\paragraph{Adversarial Loss}
The multi-scale discriminator~\cite{pix2pixHD} $D$ consists of three sub-discriminators $\{D_i\}_{i=1,2,3}$. The adversarial loss for generator $G_a$ and $D$ is defined as:
\begin{equation}
\label{eqn:new_loss_adv}
\begin{aligned}
\mathcal{L}_{adv}(G_a;D) &=\frac{1}{3}\sum_{i=1}^{3}\mathbb{E}\big[\log D_i({\synS},{x})\big] \\
& + \frac{1}{3}\sum_{i=1}^{3} \mathbb{E}\Big[\log \Big(1-D_i \big({\synS},G_a({\synS})\big)\Big)\Big].
\end{aligned}
\end{equation}
The adversarial loss for $G_m$ and $D$, denoted as $\mathcal{L}_{adv}(G_m;D)$ is defined similarly.

\paragraph{Discriminator Feature Matching Loss} Similar to pix2pixHD~\cite{pix2pixHD} and Lines2FacePhoto~\cite{Lines2Face}, we use a discriminator feature matching loss as the perceptual loss, which is designed to minimize the content difference between generated image and the real image in feature space. The discriminator feature matching loss uses the discriminator $D$ as the feature extractor. Let $D^q_i()$ be the output of $q^{th}$ layer in $D_i$. This feature matching loss is defined as:
\begin{equation}
\label{eqn:feature_matching_loss}
\begin{aligned}
\mathcal{L}_{DFM}(G_a) & =\frac{1}{3N_Q}\mathbb{E}\sum_{i=1}^{3}\sum_{q\in Q} \frac{1}{n_i^q} \|D^q_i\big({\synS},G_a({\synS})\big) \\
&-D^q_i\big({\synS},{x}\big)\|_1 ,
\end{aligned}
\end{equation}
where $Q$ is the selected layers of discriminator for computing this loss, $N_Q$ is the number of elements in $Q$, $n^q_i$ is the number of elements in $D^q_i$.
The discriminator feature matching loss for $G_m$ and $D$, denoted as $\mathcal{L}_{DFM}(G_m)$, is defined similarly.

\paragraph{Generator Feature Matching Loss}
Similar to discriminator feature matching loss which is designed to minimize the content difference between generated images and real images in feature space, our generator feature matching loss aims to minimize the content difference between the embedding of edge-aligned sketches $\synS$ and deformed sketches $\dfmS$ in the generator feature space. Let $G_a^t()$ and $G_m^t()$ be the output of $t^{th}$ layer in $G_a$ and $G_m$ respectively. This loss is calculated as:
\begin{equation}
\label{eqn:loss_GFM}
\mathcal{L}_{GFM}(G_a, G_m)=\mathbb{E}\frac{1}{N_T} \sum_{t\in T}  \frac{1}{|n^t|} \|G_a^t(\synS)-G_m^t(\dfmS) \|_1,
\end{equation}
where $T$ is the set of selected generator layers for calculating this loss, $N_T$ is the number of elements of $T$, and $n^t$ is the number of elements of $G_a^t()$ and $G_m^t()$.
We select the feature maps of the first four residual blocks of the two generators in our experiments.

The overall objective of our dual-generator model is:
\begin{equation}
\label{eqn:new_minmax_game}
\begin{aligned}
\min_G  &\max_{D} \mathcal{L}_{rec}(G_a) + \mathcal{L}_{rec}(G_m) \\
& +\mathcal{L}_{adv}(G_a, D)+ \mathcal{L}_{adv}(G_m, D) \\
& +\lambda \big(\mathcal{L}_{DFM}(G_a, D) +\mathcal{L}_{DFM}(G_m, D)\big) \\
& +\mu \mathcal{L}_{GFM}(G_a, G_m),
\end{aligned}
\end{equation}
where $\lambda$ and $\mu$ are the weights for balancing different losses. We set $\lambda=10$ and $\mu=10$ in our experiments.

In order to train our model more stably, we introduce a multi-stage training schedule. At the first stage, we use edge-aligned sketches and real images to train $G_a$ and $D$ without loss function related to $G_m$. At the second stage, we train SAP and the encoder of $G_m$ and $D$ from scratch while fixing weights of other parts. Note that the residual blocks and the decoder of $G_m$ that share weights with those of $G_a$ keep unchanged in this stage. At the last stage, we finetune the whole model with the overall objective in Eq.~\ref{eqn:new_minmax_game}. 


\section{Experiments and Discussions}
\label{sec:experiments}

Trained with synthesized deformed sketches, our method is robust to hand-drawn sketches. We conduct extensive experiments to demonstrate the effectiveness of our model in generating high-quality realistic face image from sketches drawn by different users with diverse painting skills.

\subsection{Implementation Settings}
Before showing experimental results, we first introduce details in our network implementation and training.

\paragraph{Implementation Details}
We implement our model on Pytorch. Both generators for edge-aligned sketches and deformed sketches share an encoder-residual-decoder structure with shared weights except that an SAP module is added to the front of the main generator $G_m$ for deformed sketches. 
The encoder consists of four convolutional layers with $2\times$ downsampling, while the decoder consists of four convolutional layers with $2\times$ upsampling. 
Nine residual blocks between the encoder and decoder enlarge the capacity of the generators.  
The multi-scale discriminator $D$ consists of three sub-networks for three scales separately, same as Pix2PixHD~\cite{pix2pixHD}. 
Instance normalization~\cite{IN} is applied after the convolutional layers to stabilize training. 
ReLU is used as the activation function for generators and LeakyReLU for the discriminators. 
\paragraph{Data}
To produce triplets of sketch, deform sketch, and real image, $(\synS, \dfmS, x)$ for our network training, we use 
CelebA-HQ~\cite{PGGAN}, a large-scale face image dataset which contains 30K $1024\times1024$ high-resolution face images. 
CelebAMask-HQ~\cite{CelebAMask-HQ} offers manually-annotated face semantic masks for CelebA-HQ with 19 classes including all facial components and accessories such as skin, nose, eyes, eyebrows, ears, mouth, lip, hair, hat, eyeglass, earring, necklace, neck, and cloth. We utilize semantic masks in this dataset to extract semantic boundary maps as edge-aligned sketches. 
Deformed sketches are generated by vectorizing and adding random offsets to edge-aligned sketches as discussed in the last section.
Real images, sketches, and deformed sketches are resized to $256\times256$ in our experiments.

\paragraph{Training Details}
All the networks are trained by Adam optimizer~\cite{Adam} with $\beta_1=0.5$ and $\beta_2=0.999$. 
For each training stage, the initial learning rate is set to $0.0002$ and starts to decay at the half of training procedure. 
We set batch size as 32. 
The entire training process takes about three days on four NVIDIA GTX 1080Ti GPUs with 11GB GPU memory.

\paragraph{Baseline Model} 
Pix2pixHD~\cite{pix2pixHD} is a state-of-the-art image-to-image translation model for high-resolution images. 
With the edge-aligned sketches and real face images, we train pix2pixHD with its low-resolution version of generator (`global generator') as a baseline model in our experiment, denoted as \textit{baseline}. 
In order to conduct a fair comparison on generalization, we also train the baseline model with augmented dataset by adding pairs of deformed sketches and images, denoted as \textit{baseline\_deform}.
The key idea of our method is using SAP and dual generators to improve the tolerance to sketch distortions. The local enhancer part of pix2pixHD, which is designed for high-resolution image synthesis, can be easily added to improve fine textures for both baseline models and our model in the future.

\subsection{Evaluation on Generation Quality}

\subsubsection{Evaluation Metrics} 
Evaluating the performance of generative models has been studied for a period of time in image generation literature.
It is proven to be a complicated task because a model with good performance with respect to one criterion does not necessarily imply good performance with respect to another criterion~\cite{GANs_equal}. 
A proper evaluation metrics should be able to present the joint statistics between conditional input samples and generated images.
Traditional metrics, such as pixel-wise mean-squared error, can not effectively measure the performance of generative models. 
We utilize two popular quantitative perceptual evaluation metrics based on image features extracted by DNNs: Inception Score (IS)~\cite{Improved_Techniques} and Fréchet Inception Distance (FID)~\cite{FID}.
These metrics are proven to be consistent with human perception in assessing the realism of images.

\paragraph{Inception Score (IS)}
IS applies an Inception model that is pre-trained on ImageNet to extract features of generated images and computes the KL divergence between the conditional class distribution and the marginal class distribution. Higher IS presents higher quality of generative images.
Note that IS is reported to be biased in some cases because its evaluation is based more on the recognizability rather than on the realism of generated samples~\cite{evaluation}. 

\paragraph{Fréchet Inception Distance (FID)}
FID is a recently proposed evaluation metric for generative models and proven to be more consistent with human perception in assessing the realism of generated images. FID computes Wasserstein-2 distance between features of generated images and real images which are extracted by a pre-trained Inception model. Lower FID indicates that the distribution of generated data is closer to the distribution of real samples.

\subsubsection{Image Quality Comparison}
Existing image-to-image translation models can be trained for sketch-to-face translation using paired sketch and image data. 
Since the quality of generated images presents the basic performance of a generative model, we first compare the quality of generated images by different generative models using IS and FID. 
We test these models with edge-aligned sketches that are synthesized from images in the test set. 
Besides the baseline model, we also test Pix2pix~\cite{pix2pix}, which is the first general image-to-image translation framework. It can be applied to a variety of applications by switching the training data. 
We use the default setting to train Pix2pix model with paired edge-aligned sketches and real face images. 
All these methods produce face image in dimension of $256\times256$.

\begin{figure}
	\includegraphics[width=0.9\linewidth]{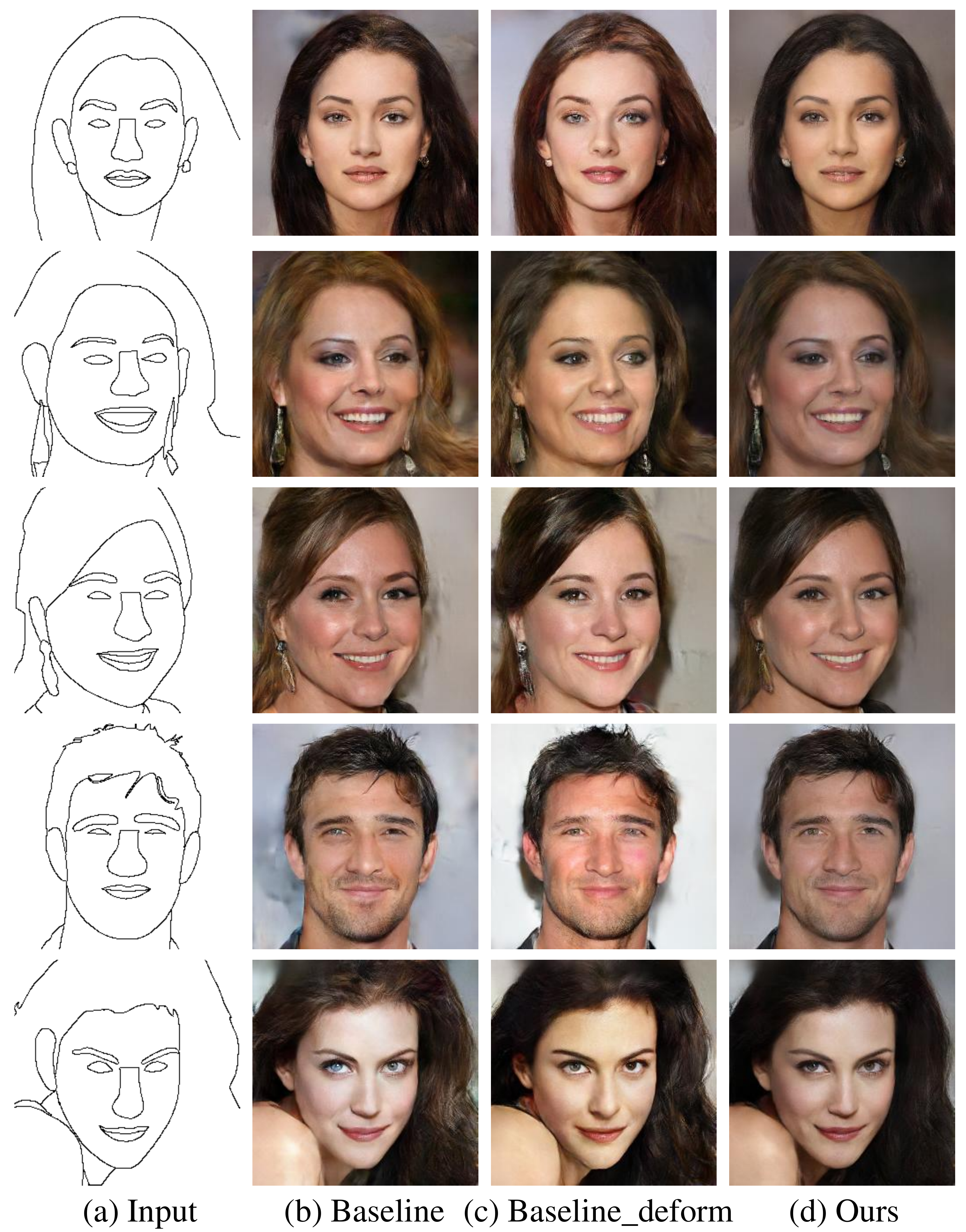}
	\caption{Both our model and existing models, which generate plausible photo-realistic face images from synthesized sketches in training set, are able to be generalized to synthesized test sketches without deformation that shows similar distribution with the training data distribution. }
	\label{fig:generative_quality}
\end{figure}

Table~\ref{tab:generative_quality} shows the quantitative evaluation results of four models. Our model surpasses other models by a small margin with respect to two evaluation metrics. 
Visual results are shown in Figure~\ref{fig:generative_quality}. As we can see, all of the four models are able to generate plausible face images from sketches of test dataset. Since these test sketches are generated using the same method as training sketches, the test data distribution is quite close to the training data distribution. Both our model and existing models perform well on these samples with ideal face structures. 
We will show the superiority of our method on more challenging sketches in the following experiments.

\begin{table}[h]
	\centering	
	\caption{Quantitative comparison of generative quality on synthesized test sketches.}
	\begin{tabular}{|c|c|c|c|c|}\hline
		& Pix2pix \cite{pix2pix} & Baseline~\cite{pix2pixHD} & Baseline\_deform & Ours \\\hline
		IS & $2.186$ & $2.298$ & $2.369$ & $\textbf{2.411}$\\\hline
		FID & $289.3$ & $259.1$ & $244.3$ & $\textbf{242.1}$\\\hline
	\end{tabular}
	\label{tab:generative_quality}
\end{table} 

\subsection{Comparison of Generalization Capability}

In order to verify the generalization ability of our model, we compare our model with state-of-the-art image translation models by testing with synthesized sketches of different levels of deformation, well-drawn sketches and poorly-drawn sketches by common users.

\subsubsection{Different Level of Deformation}
As mentioned in Section \ref{subsec:algorithm_data}, we deform an edge-aligned sketch $\synS$ to obtain a corresponding deformed sketch $\dfmS$ by adding random offsets to the control points and end points of the vectorized strokes in $\synS$. 
The maximum offset $d$ is set to $11$ in the training data. 
We further create more deformed sketches with multiple levels of deformation, denoted as $\dfmS^d$, by modifying the maximum offset $d$.
We examine the generalization ability of our model and the baseline model on these deformed sketches. 
Note that the \textit{baseline} is trained with only edge-aligned sketches, while our model and \textit{baseline\_deform} model are trained with both edge-aligned sketches and deformed sketches with $d=11$.

\begin{figure}
	\includegraphics[width=0.9\linewidth]{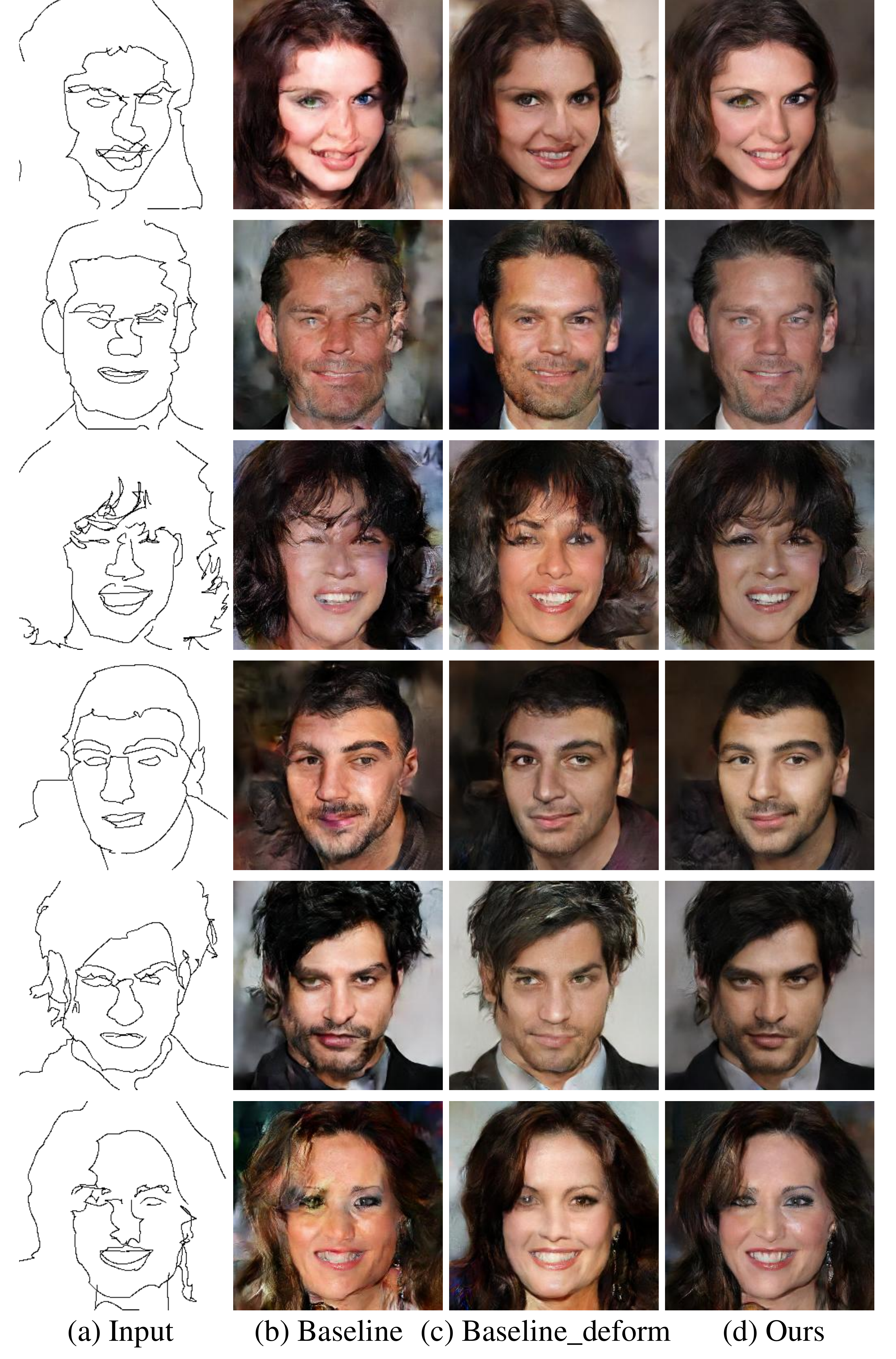}
	\caption{For sketches with large deformation, both \textit{baseline} model and \textit{baseline\_deform} model fail to generate satisfying results. Artifacts can be found in areas with large sketch deformation in (b) and (c). Our results maintain high image quality even large deformation occurs in the input sketch.  }
	\label{fig:generalization_examples}
\end{figure}

In this experiment, the input sketches are deformed by larger offsets where the maximum $d$ is set to $30$. 
As shown in Figure~\ref{fig:generalization_examples}, strokes in sketches with larger deformation looks quite different from those in the training sketches including edge-aligned sketches $\synS$ and deformed sketches $\dfmS^{11}$. 
By adding deformed sketches into training data, \textit{baseline\_deform} produces better images than \textit{baseline}. However, when larger deformation occurs, \textit{baseline\_deform} suffers from artifacts in facial features, for example, the mouth in the first example, the eyes of the third and the last case in Figure~\ref{fig:generalization_examples}, 
In comparison, our model produces more realistic face images with more symmetric eyes and fine textures, benefited from its ability of capturing the distortion of deformed strokes and rectifying shape features by our spatial attention pooling module. 

\subsubsection{Hand-Drawn Sketches}
Besides the synthesized sketches with stroke deformation, we further examine the model generalization ability by comparing performances of our model with baseline models on two kinds of hand-drawn sketches: expert-drawn sketches and common-user sketches drawn by users without professional painting skills.

\begin{figure}
	\includegraphics[width=0.9\linewidth]{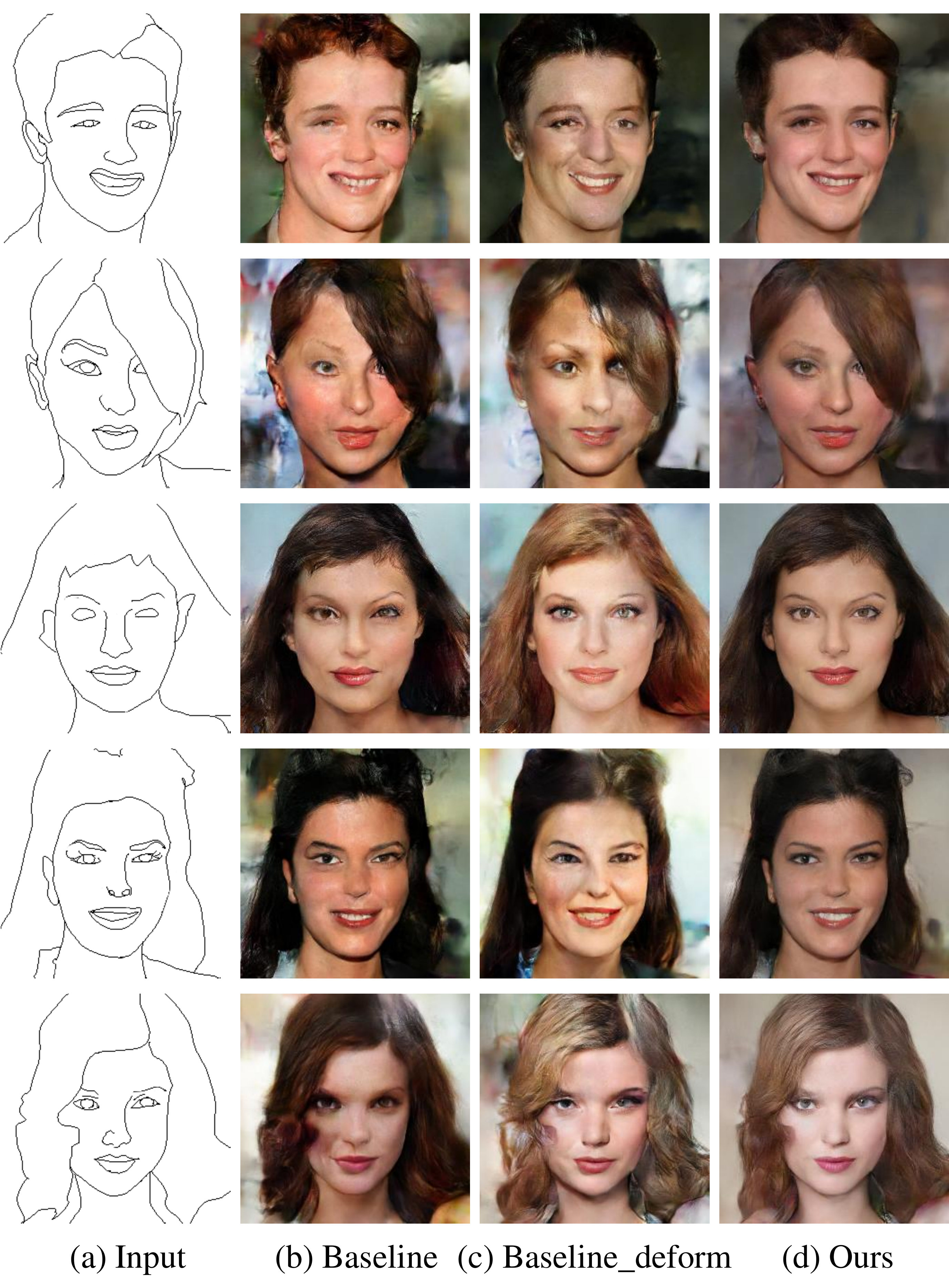}
	\caption{Our model is successfully generalized to well-drawn expert sketches, while the result quality of baseline models degenerates even trained with deformed sketches.}
	\label{fig:expert_sketches}
\end{figure}

\paragraph{Expert-Drawn Sketches}
We invite an expert with well-trained drawing skills to draw 12 portrait sketches for testing. These expert sketches were drawn on a pen tablet so that the strokes are smooth and precise. Note that shading strokes are not drawn. Figure~\ref{fig:expert_sketches} shows a group of face images generated by different models from several expert-drawn sketches. Even with well-drawn strokes, the \textit{baseline} and \textit{baseline\_deform} frequently fail to produce realistic textures and complete structures of eyes or mouths.
In comparison, our results are more realistic with fine textures and intact structures. 

\begin{figure}
	\includegraphics[width=0.9\linewidth]{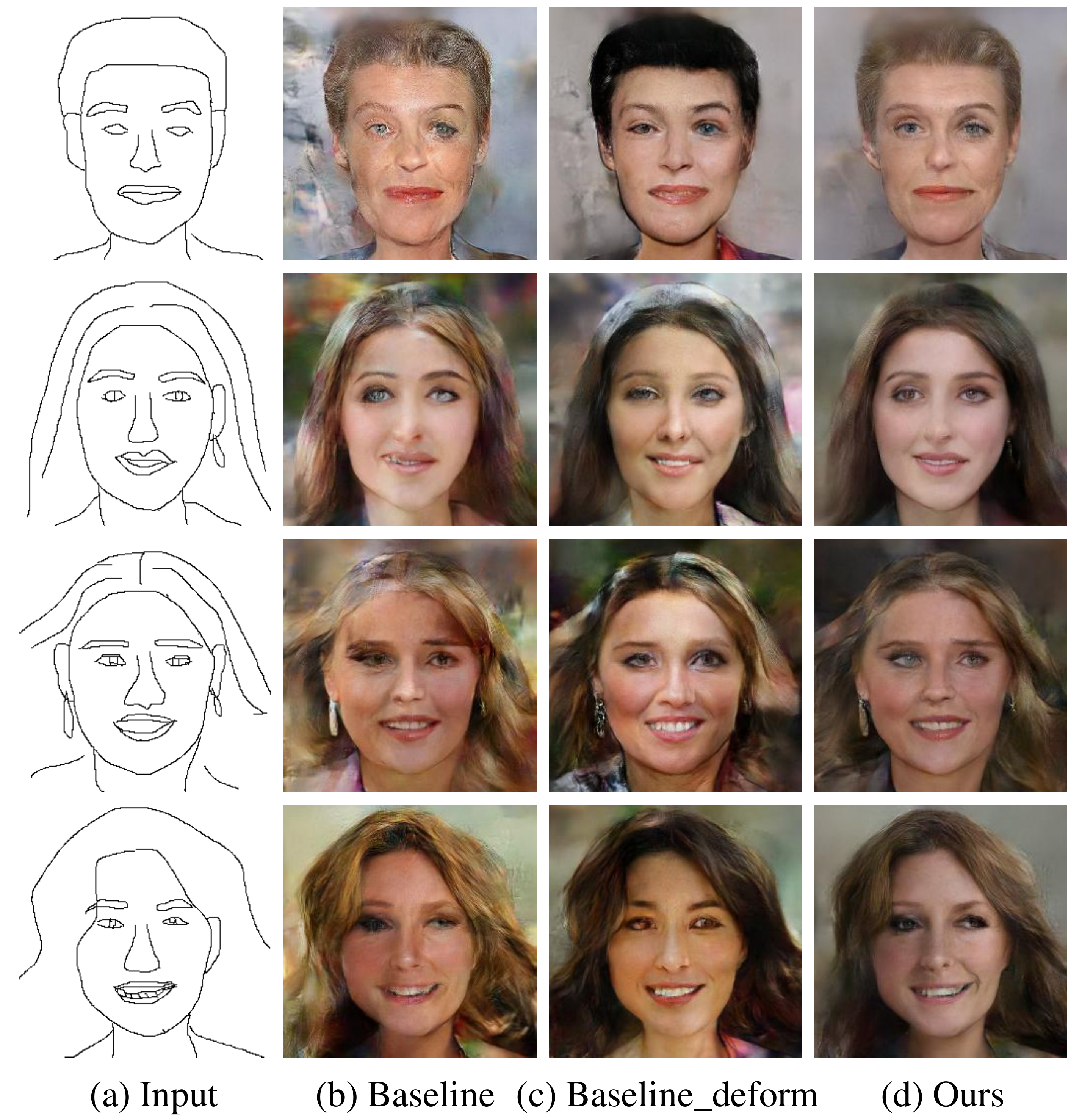}
	\caption{For these challenging sketches drawn by common users, our model is able to generate plausible results. In comparison, the results of baseline models are over blurry and present obvious artifacts in fine textures.}
	\label{fig:common_sketches}
\end{figure}

\paragraph{Common Sketches}
We also invited 20 graduate students without drawing skills to draw 200 freehand sketches of their imagined faces using mouses. Hence, strokes of these common sketches roughly depict the desired face structure and shapes of facial features with some distortion. 
Moreover, common sketches show different levels of details. For example, some sketches contains many strokes inside the hair regions, which are typically blank in the training sketches. Results shown in Figure~\ref{fig:common_sketches} demonstrate that our model is robust to these poorly-drawn sketches. 
In contrast, the diversity of stroke styles and detail levels significantly damage the visual quality of the generated results from the \textit{baseline} and \textit{baseline\_deform} models.

\section{Conclusion}
In this paper, we present DeepFacePencil, a novel deep neural network which allows common users to create photorealistic face images by freehand sketching. The robustness of our sketch-based face generator comes from the proposed dual generator training strategy and spatial attention pooling module. The proposed spatial attention pooling module adaptively adjusts the spatially varying balance between the image realism and the conformance between the input sketch and the synthesized image. By adding the SAP module to our generator and training two dual generators simultaneously, our generator effectively captures face structure and facial feature shapes from coarsely drawn sketches. 
Extensive experiments demonstrate that our DeepFacePencil successfully produces high-quality face images from freehand sketches drawn by users in diverse drawing skills.


\begin{acks}
This work was supported by the National Key Research $\&$ Development Plan of China under Grant 2016YFB1001402, the National Natural Science Foundation of China (NSFC) under Grants 61632006, U19B2038, and 61620106009, as well as the Fundamental Research Funds for the Central Universities under Grants WK3490000003 and WK2100100030. We thank the Supercomputing Center of USTC for providing computational resources.
\end{acks}

\bibliographystyle{ACM-Reference-Format}
\balance 
\bibliography{sketch}

\end{document}